\documentclass[10pt,journal,compsoc]{IEEEtran}
\ifCLASSOPTIONcompsoc
  \usepackage[nocompress]{cite}
\else
  \usepackage{cite}
\fi
\ifCLASSINFOpdf
\else
\fi
\usepackage{amsmath,amsfonts}
\usepackage{algorithmic}
\usepackage{algorithm}
\usepackage{array}
\usepackage{textcomp}
\usepackage{stfloats}
\usepackage{url}
\usepackage{verbatim}
\usepackage{graphicx}
\usepackage{cite}
\usepackage{tabu}
\usepackage{subfigure}
\usepackage{multirow}
\usepackage{booktabs}
\usepackage{threeparttable}
\usepackage{hyperref}

\begin{document}
%
\title{Cooperative Sentiment Agents for Multimodal Sentiment Analysis}
%
%
%
%

\author{Shanmin Wang,
         Hui Shuai,
         Qingshan Liu,~\IEEEmembership{Senior Member,~IEEE,}
         Fei Wang
\IEEEcompsocitemizethanks{\IEEEcompsocthanksitem
Shanmin Wang, PhD candidate in College of Computer Science and Technology in Nanjing University of Aeronautics and Astronautics, Nanjing, China. E-mail:smwang1994@nuaa.edu.cn.}
\IEEEcompsocitemizethanks{\IEEEcompsocthanksitem
Hui shuai is the assistant professor in School of Computer Science in Nanjing University of Posts and Telecommunications, Nanjing, China. E-mail: huishuai13@163.com.}
\IEEEcompsocitemizethanks{\IEEEcompsocthanksitem
Qingshan Liu, corresponding author, professor in Nanjing University of Posts and Telecommunications and PhD supervisor of Nanjing University of Aeronautics and Astronautics, Nanjing, China.
E-mail: qsliu@njupt.edu.cn.}
\IEEEcompsocitemizethanks{\IEEEcompsocthanksitem
Fei Wang, Professor in Functional Brain Imaging Institute and Nanjing Medical University, Nanjing, China. E-mail: fei.wang@yale.edu.}
\thanks{Manuscript received January 17, 2024;}}

%
%

\markboth{Journal of \LaTeX\ Class Files,~Vol.~14, No.~8, August~2015}%
{Shell \MakeLowercase{\textit{et al.}}: Bare Demo of IEEEtran.cls for Computer Society Journals}
%



\IEEEtitleabstractindextext{%
\begin{abstract}
In this paper, we propose a new Multimodal Representation Learning (MRL)  method for Multimodal Sentiment Analysis (MSA), which facilitates the adaptive interaction between modalities through Cooperative Sentiment Agents, named Co-SA. Co-SA comprises two critical components: the Sentiment Agents Establishment (SAE) phase and the Sentiment Agents Cooperation (SAC) phase. During the SAE phase, each sentiment agent deals with an unimodal signal and highlights explicit dynamic sentiment variations within the modality via the Modality-Sentiment Disentanglement (MSD) and Deep Phase Space Reconstruction (DPSR) modules. Subsequently, in the SAC phase, Co-SA meticulously designs task-specific interaction mechanisms for sentiment agents so that coordinating multimodal signals to learn the joint representation. Specifically, Co-SA equips an independent policy model for each sentiment agent that captures significant properties within the modality. These policies are optimized mutually through the unified reward adaptive to downstream tasks. Benefitting from the rewarding mechanism, Co-SA transcends the limitation of pre-defined fusion modes and adaptively captures unimodal properties for MRL in the multimodal interaction setting. To demonstrate the effectiveness of Co-SA, we apply it to address Multimodal Sentiment Analysis (MSA) and Multimodal Emotion Recognition (MER) tasks. Our comprehensive experimental results demonstrate that Co-SA excels at discovering diverse cross-modal features, encompassing both common and complementary aspects. The code can be available at https://github.com/smwanghhh/Co-SA.
\end{abstract}

}

\maketitle

\IEEEdisplaynontitleabstractindextext

%
\IEEEpeerreviewmaketitle

\IEEEraisesectionheading{\section{Introduction}\label{sec:introduction}}

\par{\IEEEPARstart{T}{he} rapid development of deep learning~\cite{lecun2015deep, he2016deep} and reinforcement learning~\cite{jiang2023variance, deng2023false} has 
propelled the field of human-computer interaction remarkably.
However, human-machine interactions are predominantly task-oriented yet lack some spontaneous and personalized service due to machines' inability to gain individuals' inner states~\cite{spezialetti2020emotion, chen2021patient, xing2023relational}.
Consequently, a refined sentiment analysis algorithm is urgently required for the new-generation intelligent human-machine interaction~\cite{minsky2007emotion, lian2023gcnet, fan2022emotional, zeng2007survey}. In daily life, people unconsciously emit their inner states through a mixture of spoken language, visual clues, and acoustic signals. 
Analyzing sentiment states or emotion labels in such multimodal contexts is known as multimodal sentiment analysis (MSA)~\cite{soleymani2017survey} or multimodal emotion recognition (MER)~\cite{sebe2005multimodal}. }

\par{Multimodal Representation Learning (MRL) integrates multiple heterogeneous signals into a joint representation and plays a crucial role in the MSA task~\cite{baltruvsaitis2018multimodal}. It has derived diverse modality-fusion strategies, including tensor-based operations~\cite{zadeh2017tensor, liu2018efficient}, Canonical Correlation Analysis (CCA)~\cite{gao2017discriminative, sun2005new}, graph-based fusion~\cite{jin2022social, zhang2021multi}, attention mechanisms~\cite{zadeh2018memory, zadeh2018multi}, etc. In recent years, self-attention and transformers have become mainstream technologies for modality fusion due to their distinctive structures in integrating multiple representations~\cite{vaswani2017attention}.
Nevertheless, within the self-attention mechanism, the inner product between the query and key, which are constructed by different modalities, highlights inter-modal similarities~\cite{tsai2019multimodal, yang2022disentangled}.
It indicates that self-attention-based models extract more modality-shared properties for constructing the joint multimodal representation. 
However, parts of modality-unique features also complement each other. Hence, there is an urgent need to transcend the modality-shared bias of predefined fusion modes, coordinate adaptive interactions among multiple signals, and achieve precise predictions in individual sentiment states. Besides, capturing sentiment variation over time from heterogeneous signals and representing them in a unified form is an important premise for modality fusion.}

\par{In this paper, we coordinate the adaptive interaction between multiple signals to learn the joint representation through Cooperative Sentiment Agents (Co-SA). Co-SA comprises the Sentiment Agent Establishment (SAE) phase and the Sentiment Agent Cooperation (SAC) phase, which capture sentiment variations within unimodal signals and foster interaction between multiple unimodal representations, respectively. Specifically, in the SAE phase, each sentiment agent contains a Modality-Sentiment Disentanglement (MSD) module and a Deep Phase Space Reconstruction (DPSR) module to express sentiment dynamics in a unified form. The MSD module disentangles sentiment features from raw input, thereby mitigating the impact of diverse modal properties. Inspired by Phase Space Reconstruction~\cite{takens1981detecting}, the DPSR module establishes relationships between short- and long-time observations, emphasizing sentiment variations over time. In the SAC phase, each sentiment agent takes actions based on the respective policy model to determine those unimodal properties that contribute to the joint representation. To further facilitate the interaction between multiple signals, Co-SA defines the task-tailored cooperation mechanism between sentiment agents in policy models and their joint optimization strategies. It also rewards all sentiment agents jointly for their actions, so as to adjust policies mutually. Through the rewarding mechanism, Co-SA will reach the optimal multimodal representation learning strategy. To validate the effectiveness of Co-SA, we conduct comprehensive experiments on multimodal sentiment analysis (MSA) and multimodal emotion recognition (MER) tasks.}

\begin{itemize}
\item We propose the concept of Cooperative Sentiment Agents for the MSA task, which coordinates adaptive interactions across multiple modalities to learn the joint representation. \\

\item Each sentiment agent deals with heterogeneous unimodal signals with the Modality-Sentiment Disentanglement (MSD) and the Deep Space Reconstruction (DPSR) modules to highlight sentiment variations over time in a unified form, which lays a solid foundation for modality interactions. \\

\item To facilitate the interaction across multiple modalities in the MRL task, we design a task-tailored collaboration mechanism for sentiment agents through policy models. These policies are further mutually optimized via a unified reward, in conjunction with the optimal joint representation achieved.\\
\end{itemize}

\begin{figure*}[t]
  \centering
    \includegraphics[width=1.0\linewidth]{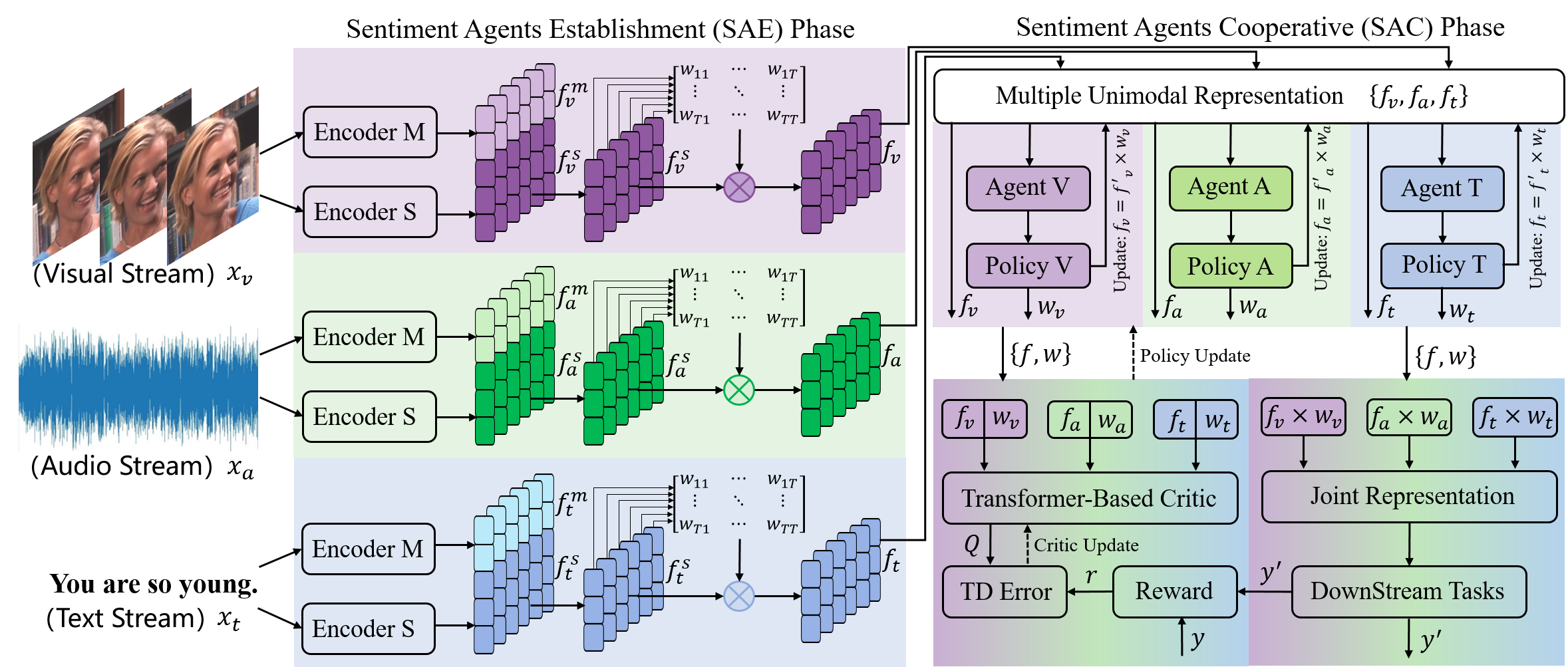}
    \caption{Framework of the Co-SA model. 'V', 'A', and 'T' represent visual, acoustic, and text modalities, respectively. 'Encoder M' and 'Encoder S' indicate modality and sentiment encoders, respectively. $f^m$ and $f^s$ are modality and sentiment features, respectively. $\otimes$ denotes the inner product between sequential features and matrix. }
    \label{fig:framework}
\end{figure*}

\section{Related Work}
\subsection{Multimodal Sentiment Analysis}
\par{Multimodal sentiment analysis (MSA) aims to infer individuals' sentiment states through the cues they inadvertently emitted. These cues encompass behavioral aspects (e.g., facial expressions) and physiological indicators (e.g., electroencephalogram). With the recent surge in the popularity of smart devices, a multitude of users create and upload videos to platforms like YouTube to share their immediate moods. This exponential growth in video data has laid a robust foundation for MSA, predominantly focusing on behavioral cues. An intrinsic difficulty for MSA is to fuse heterogeneous data collected by disparate sensors for a more reliable prediction. Based on the data integration stage, existing MSA works can be categorized into feature-level fusion, decision-level fusion, and hybrid fusion.}

\par{\textbf{Feature-level fusion} consolidates several independent representations into a joint representation and subsequently makes decisions~\cite{zadeh2017tensor}. \textbf{Decision-level fusion} independently processes the features and classifiers of each modality~\cite{lopes2021automl}. It combines the decision vectors from all classifiers to produce the ultimate sentiment prediction. \textbf{Hybrid fusion} represents a combination of both feature-level and decision-level fusion techniques~\cite{li2021cognitive, sun2021two}. Among the above paradigms, decision-level fusion is the most straightforward due to the consistent formats of decision vectors. However, achieving precise predictions through decision-level fusion remains challenging. First, it is hard to determine the reliabilities of modalities when fusing their decision vectors. 
Secondly, the independent prediction process for each modality fails to capture the relevance between multimodal signals, such as shared and complementary properties.}

\par{Unlike decision-level fusion, feature-level fusion makes the most of interacted properties across multimodal signals and combines multiple representations into a powerful joint one. It has become the most popular modality-fusion mode in the MSA task. Tensor-based, graph-based, transferred-based, and attention-based techniques are popular for feature-level fusion. \textbf{Tensor-based} works employ common tensor operations to integrate multimodal representations, like concatenation, outer product, etc. TFN (Tensor Fusion Network) uses a three-fold Cartesian product to establish trimodal dynamic explicitly~\cite{zadeh2017tensor}. LMF (Low-rank Multimodal Fusion) analyzes that the TFN, performing outer product on high-dimension embedding, has resulted in the computational burden. 
To optimize TFN, it exploits the parallel decomposition of low-rank weight tensor and input tensor to compute tensor-based fusion~\cite{liu2018efficient}. Analogously, MRRF uses Tuckers' tensor decomposition to disentangle unimodal representations and remove redundant parts to reduce computation. It gets joint multimodal representations by performing the outer product on disentangled modal embeddings~\cite{fung2019modality}. Some works also construct \textbf{graphs} to learn the joint embedding, where nodes and edges represent modalities and corresponding relationships~\cite{jin2022social}. HHMPN successively models the feature-to-label, label-to-label, and modality-to-label dependencies via passing graph messages~\cite{zhang2021multi}.
\textbf{Transfer-based} technology is another popular way of learning the joint representation by transferring one modal representation into another with the encoder-decoder structure. The immediate features output by the encoder are regarded as the joint representation of the two modalities~\cite{mai2020modality}. The joint representation for all modalities will be obtained by transferring signals in succession. 
Some works avoid the sequential transfer among modalities. They specify the dominant modality in advance and then transfer other modalities to the specified one to strengthen it. The \textbf{attention} mechanism is widely used for feature-level fusion since it inspires models to focus on some vital features when learning the joint representation.
MFN proposes the Delta-memory Attention Network (DMAN) to learn fusion weights for disparate multimodal elements through time. It also designs the Multi-view Gated Memory module to filter noisy features~\cite{zadeh2018memory}. MARN applies the attention mechanism to the hidden state of LSTM and aggregates multimodal representations~\cite{zadeh2018multi}. MulT applies the intra-modal transformer and cross-modal transformers to alleviate the issue of (1) unaligned multimodal data; (2) long-time dependencies between elements across modalities~\cite{tsai2019multimodal}. GME-LSTM(A)~\cite{chen2017multimodal} is devoted to addressing unimodal temporal dynamics and multimodal importance issues. It applies temporal attention to the output of the LSTM module to attend to the vital time steps and modal attention to emphasize the significance of each modality.}

\par{Although the above feature-level techniques have successfully merged multiple signals into a joint representation in form, they care more about modality-fusion strategies yet lack deliberation about specific features suitable for fusion. Recently, there has been a growing awareness that diverse inter-modal properties may significantly contribute to the joint multimodal representation. Several studies have divided raw signals into modality-shared and modality-unique representations prior to the fusion process~\cite{li2023decoupled, yang2022disentangled, harada2017deep, he2022dynamic}. However, these methods still struggle with the challenge of capturing appropriate features from shared and unique components to accurately predict sentiment states. A novel technique that can overcome the limitations of existing fusion modes and adaptively capture significant properties from unimodal signals for the joint representation is still needed. In this paper, we propose to collaboratively deal with multiple signals to learn the joint representation through cooperative sentiment agents.}

\par{Except for the modality-fusion technique, the quality of unimodal representations also significantly impacts the MSA task. Firstly, multimodal data are collected by disparate sensors and represented using distinct digital signal formats. Video, audio, and text data, for instance, are recorded via frames, waveforms of amplitude values over time, and sets of characters, respectively. Meanwhile, each modality carries different levels of modal and sentiment properties. It is essential for MRL to disentangle sentiment features from heterogenous data and represent them in a unified format. Secondly, features from adjacent frames often exhibit severe overlap and minor value fluctuations, making them weak at reflecting temporal dynamics. Moreover, prevalent approaches typically aggregate multimodal representations initially and subsequently extract temporal features, inadvertently neglecting the exploration of unimodal sentiment variations over time. The heterogeneous multimodal data and implicit temporal dynamics have increased the difficulty for sentiment agents in learning significant cross-modal features. To alleviate the above issues, we propose the Modality-Sentiment Disentanglement (MSD) and Deep Phase Space Recognition (DPSR) modules when establishing sentiment agents.}

\subsection{Multi-Agent Cooperative Reinforcement Learning}

\par{Reinforcement Learning (RL)~\cite{kaelbling1996reinforcement} is about an agent spontaneously interacting with the environment, taking actions according to policies, and rewarding them. The environmental state $s$, agent's policy $\pi$, state transition $p$, and reward function $r$ are essential components of RL. Supposing at time $t$, the agent perceives the state $s^t$ from the environment and takes action $a^t$ based on the policy $\pi ^ t$. Upon taking the action, the agent receives a reward $r^t$, and the environmental state is updated to $s^{t+1}$. The agent repeats the above process until the terminal state $s^{t+k}$, with a sequence of rewards $\{ {r^t}, \cdots ,{r^{t + k}}\}$ accrued. The key to RL is to learn the optimal policy that maximizes the cumulative reward. The Actor-Critic model is a popular algorithm for RL, in which the actor and critic models are used to learn and evaluate policies, respectively~\cite{kober2013reinforcement}. Cooperative multi-agent reinforcement learning involves multiple agents working together to achieve a coincident goal~\cite{tampuu2017multiagent, kok2006collaborative}. Centralized learning with fully decentralized execution is one of the prominent approaches for cooperative agents~\cite{foerster2018counterfactual, rashid2020monotonic}. It entails a central control unit that observes agents' states and regulates their actions during training. After the training phase, the central unit is removed, and each agent independently makes decisions based on the trained policies.}

\section{Multimodal Representation Learning Based on Cooperative Sentiment Agents}
\par{In this paper, we propose an exclusive sentiment agent for each modality and further collaboratively control multiple sentiment agents to promote multimodal interactions and learn the joint multimodal representation. As shown in Fig.~\ref{fig:framework}, the proposed framework comprises two essential components: the Sentiment Agents Establishment (SAE) Phase and the Sentiment Agents Cooperative (SAC) Phase. 
In the SAE Phase, sentiment agents disentangle sentimental properties from heterogeneous signals to the same form through the MSD module. They further contain the DPSR module that ferrets out sentiment sequential variations to obtain high-quality unimodal representations. In the SAC Phase, Co-SA collaboratively controls sentiment agents to learn the joint representation, where each sentiment agent determines effective features within its modality to contribute to the joint representation.
Co-SA further applies the joint representation to predict downstream tasks and subsequently calculates unified rewards to guide sentiment agents in adjusting their policies. In the following section, we will provide a detailed introduction to the SAE and SAC Phases, respectively.
}

\subsection{Sentiment Agents Establishment}
\par{Individuals express their sentiment states through multiple behavioral signals, which are collected and recorded by various devices. The elements comprising each modal signal vary across modalities. For instance, text signals consist of sets of characters, while visual data is a collection of frames. Additionally, different modalities exhibit substantial variations in modal properties, noise, and even task relevance. These significant differences pose challenges for sentiment agents when uniformly dealing with multimodal signals during the modality-fusion process. Meanwhile, diverse modal properties also hinder sentiment agents from precisely extracting sentiment-related features. To provide precise sentiment signals for each sentiment agent, we propose the Modality-Sentiment Disentanglement (MSD) module to extract sentiment-related properties during the representation fission process.}
\subsubsection{Modality-Sentiment Disentanglement (MSD)}
\par{Each modality carries both sentiment-related and modality-specific properties in different forms. Consequently, we design a sentiment encoder $E_i^s$ and a modality encoder $E_i^m$ to extract sentiment-related $f_i^s$ and modality-specific $f_i^m$ features from raw input $x_i$, respectively, where $i$ can be selected text $t$, visual $v$, and acoustic $a$ data.

\begin{equation}
f_i^s = E_i^s(x_i;\theta _i^s)
\label{equ:dis_s}
\end{equation}

\begin{equation}
f_i^m = E_i^m(x_i;\theta _i^m)
\label{equ:dis_m}
\end{equation}
where $\theta_i^s$ and $\theta_i^m$ are parameters of sentiment and modality encoders, respectively. Features extracted by the modality encoder vary across modalities. To facilitate the modality encoder in extracting as many precise modal properties as possible, we constrain $f_i^m$ as follows.}
\begin{equation}
  {L_m} = -\frac{1}{3}\sum\limits_i^{\{ v,a,t\} } {{y_i}\log {F^m}(f_i^m;{\theta ^m})} 
  \label{equ:3}
\end{equation}
where $F^m$ indicates the function that maps $f_i^m$ to predicted probabilities. Without the loss of generality, $y_v$, $y_a$, and $y_t$ are set as 0, 1, 2, respectively.
Moreover, $f_i^m$ and $f_i^s$ represent two distinct properties within the same modality. Therefore, to enable $f_i^s$ to express appropriate sentiment features and prevent the inclusion of modal properties, we impose a significant difference between $f_i^m$ and $f_i^s$.
\begin{equation}
    {L_c} =  - \frac{1}{3}\sum\limits_i^{\{ v,a,t\} } {d(f_i^s,f_i^m)}
    \label{equ:difference}
\end{equation}
Where $d$ is the euclidean distance. 
$L_c$ prevents $f_i^s$ from conveying any modality-related properties. Additionally, we decouple each modal signal into modality-related and sentiment-related information. To ensure the completeness of disentangled features, we combine the two parts and reconstruct the input ${\tilde x}_i$, preventing the loss of information.

\begin{equation}
    {L_r} = \frac{1}{3}\sum\limits_i^{\{ v,a,t\} } {d({x_i},{{\tilde x}_i})}
    \label{equ:recover_loss}    
\end{equation}
where
\begin{equation}
{{\tilde x}_i} = {D_i}(f_i^s \oplus f_i^m;{\theta ^d _i})
\label{equ:recover}
\end{equation}

where $D_i$ and $\theta _i ^d$ are the decoders and their parameters, respectively. $\oplus$ indicates the concatenation operation. $\tilde x_i$ is reconstructed unimodal signal. $d$ is the euclidean distance. 

\par{To sum up, the MSD module has three objective functions as follows.
\begin{equation}
    {L_{msd}} = {L_m} + {L_r} + {L_c}
    \label{equ:7}
\end{equation}
The modality loss $L_m$ allows each modality encoder to precisely extract modality-related properties. The reconstructed loss $L_r$ ensures the completeness of information during the disentangled process. The modality-sentiment constraint loss $L_c$ prevents the extracted sentiment embedding from carrying any modality-related properties. The stringent constraints imposed by the above objective functions ensure that sentiment agents perceive precise sentiment properties.}

\par{In addition to heterogeneous multimodal data, sentiment variation over time is also an easily overlooked issue. Most MRL models fuse unimodal signals before extracting temporal features. The assumption behind this operation is that sentiment changes in unimodal representation are well presented. However, adjacent frames often exhibit severe redundancy in semantic context, and sentiment variations only manifest in a subtle number of numerical changes, failing to explicitly capture sentiment variations. To alleviate this issue, we draw inspiration from Phase Space Reconstruction (PSR) and propose the Deep Phase Space Reconstruction (DPSR) module to recover sentiment dynamics from existing observations.}

\subsubsection{Deep Phase Space Reconstruction (DPSR)}
\par{The phase space encompasses all feasible states of a dynamical system. In this space, each point, also termed an observation, can be seen as a sampling from the motion trajectory~\cite{takens1981detecting}. These observations exhibit strong correlations, allowing us to recover the system's motions from existing observations. Simultaneously, sentiment changes in each modality can be treated as a distinct dynamic system, with each frame's representation serving as a sampled observation of the motion process~\cite{wang2020phase}. Therefore, we can apply the PSR theory to recover explicit sentiment dynamics from existing representations. Traditional PSR technology assesses the correlation between successive frames to determine time delay and embedded dimension parameters~\cite{abarbanel1993analysis, casdagli1991state}. By adjusting observations with these parameters, the phase space significantly reduces redundancy while maintaining relevance at a certain level. Building upon conventional PSR techniques but tailored to the task, the proposed DPSR module reconstructs each observation by incorporating deep learning principles to establish local and global motion information.
More precisely, the DPSR module initiates its exploration of motion information of observations by utilizing a cross-correlation matrix $W_i$ first.}

\begin{equation}
{W_i} = f_i^s{(f_i^s)^{\rm{T}}}
\label{equ:8}
\end{equation}
where $W_i$ is a $T \times T$ matrix, $T$ represents the length of sequence, and each element ${w_{pq}}$ in $W_i$ indicates the correlation between $p$ and $q$ observations. 
Subsequently, the DPSR module reconstructs each observation with the cross-correlation matrix to recover the system motion trajectory.
\begin{equation}
f_i = {W_i}f_i^s
\label{equ:9}
\end{equation}

\par{The aforementioned operations have provided a viable solution for recovering sentiment dynamics by capturing both local and global motion information. To further emphasize temporal dynamics while mitigating redundancy between frames, the DPSR module imposes constraints on the reconstructed observations as follows.

\begin{equation}
{L_{dpsr}} = \frac{1}{{T(T - 1)}}\sum\limits_p^T {\sum\limits_{q \ne p}^T {(\frac{{{{(f_i)}_p}{{(f_i)}_q}}}{{\left\| {{{(f_i)}_p}} \right\|\left\| {{{(f_i)}_q}} \right\|}}} }  + 1)/2
\label{equ:10}
\end{equation}

The objective function mentioned above aims to reduce redundancy among arbitrary pairs of frames to highlight sentiment changes. However, in some scenarios, specific events may repeat at intervals. To enhance precision in constructing sentiment motion, the DPSR module further refines the loss function as follows.

\begin{equation}
{L_{dpsr}} = \frac{1}{{T(T - 1)}}\sum\limits_p^T {\sum\limits_{q \ne p}^T {{\eta _{pq}}(\frac{{{{(f_i)}_p}{{(f_i)}_q}}}{{\left\| {{{(f_i)}_p}} \right\|\left\| {{{(f_i)}_q}} \right\|}}} }  + 1)/2
\label{equ:11}
\end{equation}
where $\eta _{pq}$ indicates the time interval between $p$ and $q$ frames, which is calculated as follows.
\begin{equation}
{\eta _{pq}} = T - \left| {p - q} \right|
\label{equ:12}
\end{equation}
Clearly, ${\eta _{pq}}$ will be small if the $p$ frame is far from the $q$ frame, and vice versa. 
}

\subsection{Cooperative Sentiment Agents for Multimodal Representation Learning}

\par{Benefitting from the SAE phase, each agent perceives unimodal representations that carry sequential sentiment variations in a unified form. These sentiment properties can either be homogeneous or complementary across modalities, often referred to as modality-shared and modality-unique features. Solely extracting shared sentiment properties from each modality can be redundant, while constructing the joint representation with only complementary parts may lead to the loss of common characteristics. Thus, extracting appropriate homogeneous and complementary sentiment properties from each unimodal representation is crucial for constituting the joint representation. Yet, it also poses great challenges to related technologies in feature learning simultaneously. Cooperative control pre-defines the communication mode between agents according to the task and adjusts to the optimal policies through rewards. This working mechanism aligns well with our goal of coordinating multiple signals to achieve the optimal joint representation. Inspired by cooperative control, we propose the concept of sentiment agents and realize the interaction between multimodal signals through agents' collaborations to finish the MRL task.}

\par{We assign sentiment agents for each modality, where unimodal representation $f_i$ is the input for each agent. Each sentiment agent has an independent policy model, also referred to as the actor model. It takes actions $w_i$ according to the current policy $\pi _i$ to determine valuable properties within the modality. Then, Co-SA combines selected features from all modalities into $f = ({f_v} \times {w_v}) * ({f_a} \times {w_a}) * ({f_t} \times {w_t})$ and further applies it to downstream tasks, where $*$ indicates the integrated mode. The key to the SAC phase is exactly hitting features that enhance or complement across modalities through $w_i$, avoiding the loss or redundancy of sentiment properties.}

\par{To this end, we carefully designed policy models and their joint optimization strategy to facilitate efficient interactions across modalities. On the one hand, we allow each sentiment agent to perceive its dominant modality as well as its differential features with other modalities. On the other hand, we jointly evaluate all policy models to coordinate captured sentiment properties across modalities. Specifically, in each stage, policy models take their dominant modal representations and different features with other modalities as input for learning common and complementary components, respectively. Taking the visual modality as an example, the visual policy model takes visual representation $f_v$ as well as its differential features with the other two modalities $f_{va}$, $f_{vt}$ as input and output the weight $W_v$.
\begin{equation}
{W_v} = {F_{\pi v}}({f_v},{f_v}_a,{f_v}_t;{\theta _{\pi v}})
\end{equation}
Where ${f_{va}} = {F_t}({f_v} - {f_a};{\theta _t})$ and ${f_{vt}} = {F_t}({f_v} - {f_t};{\theta _t})$ are differential features. Co-SA concatenates $f_v$, $f_{va}$, and $f_{vt}$ and then map to get action $w_v$. Policy models are required to be efficient on all samples. Thereby, upon outputting weights for the current batch of samples, sentiment agents transfer to the next batch of representations ${f_i} = {f_i}^\prime  \times {w_i}$ and exploit cumulative rewards to evaluate policy models. During the representation transition, ${f_i}^\prime$ is determined by the sampling order of the training samples. Meanwhile, Co-SA evaluates policy models according to the succession of weights they learned. It designs transformer-based critic models that learn cumulative rewards from the concatenation between current representations $f_i$ and weights $w_i$, avoiding the massive computation from the current to terminal training batches. Notably, all agents struggle to adjust policies and reach the consistent prediction. Thereby, Co-SA designs a joint critic model to output a unified cumulative reward for all policy models $Q = {F_c}(\sum\limits_i^{\{ v,a,t\} } {{f_i} \oplus {w_i}} ;{\theta _c})$. The joint critic model works as the central control unit that coordinates sentiment agents in continuously adjusting policies and learning appropriate cross-modal features for the joint representation. It will be removed after training, leaving policy models to decide valuable properties independently.}

\par{Clearly, policy models wish to output weights that cause a larger cumulative reward $Q$. Thereby, the objective function of policy models $L_{actor}$ is optimizing their parameters ${\theta _\pi }$ that maximize the output of the critic model. As for the optimization of the critic model, the Temporal-Difference (TD) Error algorithm~\cite{mousavi2018deep} is employed: ${L_{critic}} = Q{\rm{ - }}\bar Q$, where $\bar Q = r + \gamma Q'$. $Q'$ is the cumulative reward in the next stage, and $\gamma$ is the discounted factor. $r$ is the immediate reward, and it varies with tasks. For the MSA task, $r =  - \left| {y' - y} \right|$, where $y'$ and $y$ are predicted and true sentiment state. For the MER task, $r = \frac{{{e^{{y_i}}}}}{{\sum\limits_j^{\rm{C}} {{e^{{y_j}}}} }}$, where $y_i$ is the predicted probability for the true class, and $C$ is the number of emotion categories. Besides, the calculation of the reward $r$ relies on both the actor-critic and prediction models. To train the model stably, Co-SA iteratively optimizes the prediction and actor-critic model parameters. Thus, Co-SA contains loss functions about the policy, critic, and prediction models in the SAC phase. Co-SA has the total objective function after incorporating losses in the SAE phase.

\begin{equation}
    L = {\alpha _1}{L_p} + {\alpha _2}{L_{msd}} + {\alpha _3}{L_{dpsr}} + {\alpha _4}({L_{actor}} + {L_{critic}})
    \label{equ:21}
\end{equation}

$L_p$ is the predictive loss. It is the mean absolute loss for the MSA task and cross-entropy loss for the MER task. The weights of the four objective functions will be further discussed in the experimental part.}

\section{Experiments}
In this section, we evaluate Co-SA on multiple MRL-related tasks, i.e., multimodal sentiment analysis (MSA) and multimodal emotion recognition (MER). 

\subsection{Experimental Setting}
\subsubsection{Experimental Target}
We focus on the following questions in the experimental part: (1) whether all proposed modules are effective as expected; (2) after learning significant representation for each modality, whether it is applicable for some classic modality fusion methods; (3) If the combination of Co-SA and classic fusion methods can outperform other MSA and MER works; (4) how much of a role each modality play during the representations learning or prediction process;

\subsubsection{Databases}
\par{We apply the widely used CMU-MOSI~\cite{zadeh2016multimodal} and CMU-MOSEI~\cite{zadeh2018multimodal} databases for MSA task, and IEMOCAP~\cite{busso2008iemocap} for the MER task. CMU-MOSI, a popular database with three modalities, is widely used in MSA tasks. It is a collection of 93 videos from online sharing websites. Each video can be split into 62 utterance-level segments at most. These segments are annotated with the sentiment from [-3, 3], where -3 indicates the strongest negative sentiment while +3 represents the strongest positive attitude. Following popular works, we train Co-SA with 1281 utterances, valid, and test it with 229 and 685 utterances, respectively. CMU-MOSEI contains 2928 videos and is annotated with two types of labels, including continuous sentiment from -3 to 3 and emotions with six values. As popular MSA works, we use 16265 utterances as training data, 1869 as validation data, and 4643 as test data. IEMOCAP is the database collected in the lab, conveying three modalities of signals. It has 151 sessions of dialogues, of which there are two speakers per session and 302 videos in total. These videos are segmented into about 10 K utterances and labeled with nine emotions, including angry, happy, sad, neutral, surprised, fearful, excited, frustrated, and other. Following popular works, we take the first four emotions for experiments.
}

\subsubsection{Features Extraction}
\par{We use the following tools to extract features for three modalities.}

\par{\textbf{Visual modality.} Facet is used to extract some visual features, including action units, facial landmarks, head pose, etc., for both MSA and MER tasks.}

\par{\textbf{Acoustic modality.}
COVAREP~\cite{degottex2014covarep} is used to extract some acoustic features, such as 12-Mel frequency cepstral coefficients, pitch tracking, spectral envelope, speech polarity, etc.
}

\par{\textbf{Text modality.} Following the state-of-the-art works~\cite{rahman2020integrating, hasan2021humor, mai2023excavating, tsai2019multimodal, mai2021analyzing}, we use BERT~\cite{devlin2018bert} to extract text features for the MSA task and extract GloVe word embedding~\cite{pennington2014glove} from the original text for the MER task. }

\par{To sum up, for the \href{https://github.com/WasifurRahman/BERT_multimodal_transformer}{CMU-MOSI} database, we get the dimensions for the text, acoustic, and visual features to be 768, 74, and 47, respectively. For the \href{https://github.com/WasifurRahman/BERT_multimodal_transformer}{CMU-MOSEI} database, the feature dimensions for the three modalities are 768, 74, and 35, respectively. For the \href{https://github.com/yaohungt/Multimodal-Transformer}{IEMOCAP} database, the three modalities have respective feature dimensions of  300, 74, and 35.}

\subsubsection{Evaluations metrics} 
\par{For the MSA task, we evaluate Co-SA with (1) MAE: mean absolute error (the lower, the better); (2) Corr: correlation between predictions and ground truth (the higher, the better); (3) Acc2: binary accuracy, samples to be positive if its sentiment value is greater than 0, and vice versa; (4) F1 score; (5) Acc7: 7-class accuracy, we round up the predicted sentiment value for each sample as the class. For the MER task, we report the accuracy and F1 score for each category.}

\subsubsection{Compared Methods}
\par{(1) Tensor Fusion Network (TFN), which learns unimodal embeddings and then leverages the outer product to fuse unimodal, bimodal, and trimodal representations~\cite{zadeh2017tensor}.
(2) Low-Rank Modality Fusion (LMF), which reduces computation complexity during fusing modalities with low-rank weight tensors~\cite{liu2018efficient}.
(3) Memory Fusion Network (MFN), which designs the delta-memory attention network for learning fusion weights and multi-view gated memory to filter noisy features~\cite{zadeh2018memory}.
(4) Multimodal Transformer (MulT), which learns multimodal representations by transferring source modality to target modalities through transformers~\cite{tsai2019multimodal}.
(5) Quantum-Inspired Multimodal Fusion (QMF), which draws inspiration from the quantum principle to address the issue of multimodal combination~\cite{li2021quantum}.
(6) Graph Fusion Network (GFN), which designs graph neural networks to learn unimodal, bimodal, and trimodal dynamics~\cite{mai2020modality}.
(7) Interaction Canonical Correlation Network (ICCN), which first fuses text embeddings with visual and acoustic representations, respectively, and then uses a canonical correlation network to get unified multimodal representations~\cite{sun2020learning}. 
(8) Recurrent Attend Variation Embedding Network (RAVEN), which obtains multimodal representations by learning textual shift embeddings from visual and acoustic modalities~\cite{wang2019words}.
(9) Multimodal Adaptation Gate (MAG), which finetunes transformer model to learn efficient textural embeddings and designs multimodal gate adaptation to get multimodal representations~\cite{rahman2020integrating}.
(10) Transformer-Based Feature Reconstruction Network (TFR), which proposes a feature reconstruction network to improve the robustness of models when missing some modal signals~\cite{yuan2021transformer}.
(11) Hybrid Contrastive Learning (HyCon), which establishes complex inter-class and intra-/inter-modal relationships with contrastive and semi-contrastive learning to explore discriminative representations~\cite{mai2022hybrid}.
(12) Multimodal Correlation Learning (MCL), which learns complex correlations between modalities using prior information across samples~\cite{mai2023excavating}.
(13) Hierarchical Feature Fusion Network (HFFN), which applies the mechanism of "Divide, Conquer, and Combine" to fuse local and global multimodal representations~\cite{mai2019divide}.
(14) Temporal Convolutional Multimodal LSTM (TCM-LSTM), which leverages temporal convolutions to explore sufficient unimodal embeddings and applies the LSTM module to get multimodal representations~\cite{mai2021analyzing}.}

\begin{table*}[t]
    \centering
    \tabcolsep=0.18cm
    \setlength\extrarowheight{5pt}
    \caption{Ablation study on the MOSI and IEMOCAP databases.}
    \label{tab:ablation1}
    \scriptsize
    \begin{threeparttable}
    \begin{tabular}{cccccccccccccccc}
    \toprule[2pt]
    \multirow{3}{*}{Settings} & \multicolumn{5}{c}{Dataset: MOSI} & \multicolumn{10}{c}{Dataset: IEMOCAP} \\

    \cmidrule(r){2-6} \cmidrule(r){7-16}
    & \multirow{2}{*}{Acc7 ($\uparrow$)} & \multirow{2}{*}{Acc2 ($\uparrow$)} & \multirow{2}{*}{F1 ($\uparrow$)} & \multirow{2}{*}{MAE ($\downarrow$)} & \multirow{2}{*}{Corr ($\uparrow$) } & \multicolumn{2}{c}{Happy} & \multicolumn{2}{c}{Sad} & \multicolumn{2}{c}{Angry} & \multicolumn{2}{c}{Neutral} & \multicolumn{2}{c}{Average} \\
    \cmidrule(r){7-8} \cmidrule(r){9-10} \cmidrule(r){11-12} \cmidrule(r){13-14} \cmidrule(r){15-16}
    & & & & & & Acc & F1 & Acc & F1 & Acc & F1 & Acc & F1 & Acc & F1 \\

    \midrule[1pt]
    baseline              & \underline{\underline{47.30}} & \underline{\underline{84.88}} & \underline{\underline{84.87}} & \underline{\underline{0.711}} & \underline{\underline{0.8014}} & \underline{\underline{86.67}} & \underline{\underline{85.45}} & \underline{\underline{84.65}} & \underline{\underline{84.10}} & \underline{\underline{87.21}} & \underline{\underline{86.98}} & \underline{\underline{70.26}} & \underline{\underline{69.96}} & \underline{\underline{82.20}} & \underline{\underline{81.62}}\\
    only MSD            & \underline{\underline{48.47}} & \underline{\underline{85.50}} & \underline{\underline{85.44}} & \underline{\underline{0.711}} & \underline{\underline{0.8011}} & \underline{\underline{87.53}} & \underline{\underline{86.11}} & \underline{\underline{86.25}} & \underline{\underline{84.87}} & \underline{\underline{87.53}} & \underline{\underline{87.38}} & \underline{\underline{70.15}} & \underline{\underline{69.40}} & \underline{\underline{82.86}} & \underline{\underline{81.94}} \\
    MSD w/o $L_r$  & 47.30 & 84.88 & 84.91 & 0.711 & 0.8033 & 86.78 & 85.00 & 83.90 & 83.51 & 87.21 & 87.21 & 69.83 & 69.27 & 81.93 & 81.25 \\
    MSD w/o $L_m$  & 47.59 & 85.19 & 84.98 & 0.706 & 0.8038 & 86.89 & 85.34 & 83.16 & 82.21 & 86.67 & 86.49 & 69.51 & 69.04 & 81.56 & 80.77 \\
    MSD w/o $L_c$  & 47.74 & 85.19 & 85.06 & 0.702 & 0.8038 & 86.78 & 85.61 & 84.65 & 86.84 & 86.99 & 87.03 & 70.04 & 69.80 & 82.12 & 81.57 \\
only DPSR           & \underline{\underline{48.90}} & \underline{\underline{85.80}} & \underline{\underline{85.73}} & \underline{\underline{0.698}} & \underline{\underline{0.8055}} & \underline{\underline{87.53}} & \underline{\underline{85.72}} & \underline{\underline{86.14}} & \underline{\underline{85.82}} & \underline{\underline{88.17}} & \underline{\underline{88.45}} & \underline{\underline{72.60}} & \underline{\underline{71.82}} & \underline{\underline{83.61}} & \underline{\underline{82.95}} \\
    DPSR w/o $\eta$& 47.44 & 85.80 & 85.73 & 0.718 & 0.7947 & 86.78 & 85.67 & 85.82 & 85.45 & 87.31 & 87.20 & 70.47 & 70.10 & 82.60 & 82.11 \\
    only SAC                   & \underline{\underline{48.76}} & \underline{\underline{86.26}} & \underline{\underline{86.25}} & \underline{\underline{0.698}} & \underline{\underline{0.8006}} & \underline{\underline{87.95}} & \underline{\underline{86.24}} & \underline{\underline{86.25}} & \underline{\underline{85.63}} & \underline{\underline{87.53}} & \underline{\underline{87.60}} & \underline{\underline{71.75}} & \underline{\underline{70.34}} & \underline{\underline{83.37}} & \underline{\underline{82.45}} \\
    \bottomrule[2pt]
    \end{tabular}
    \begin{tablenotes}
    \scriptsize
    \item[] SAE and SAC represent the establishment and cooperation phases of sentiment agents, respectively. The SAE phase includes the MSD and DPSR modules. The MSD module contains the reconstructed $L_r$, modality $L_m$, and modality-sentiment constraint $L_c$ losses. The DPSR module has the weight $\eta$ in adjusting the loss function. The results from the baseline model, models incorporating only the MSD and DPSR modules, and models utilizing only the SAC phase are underlined to facilitate intuitive comparisons.
    \end{tablenotes}

\end{threeparttable}    
\end{table*}

\subsubsection{Implementation Details}
Following some state-of-the-art works~\cite{tsai2019multimodal, mai2023excavating}, Co-SA takes 50 and 20 frames of features as input for MSA and MER tasks, respectively. In the MSD module, Co-SA encodes all modal signals into 128-dimension representations for both tasks. Co-SA applies popular modality fusion modes to get the joint representation to validate its generality, including addition and concatenation. The discounted factor $\gamma$ is set as 0.5. For the weights $\alpha_1$, $\alpha_2$, $\alpha_3$ and $\alpha_4$ of loss functions, we will conduct further ablation studies on both MER and MSA tasks. Before training Co-SA, we initialize a replay memory to store a batch of features, weights, updated features, and rewards in each stage. They are sampled to optimize the actor and critic models during each training process. After optimizing the actor $\theta_{actor}$ and critic $\theta_{critic}$ models, the parameters for the target actor $\theta '_{actor}$ and critic $\theta '_{critic}$ models are updated for stable convergence: ${\theta '_{actor}} = \zeta {\theta _{actor}} + (1 - \zeta ){\theta '_{actor}}$, ${\theta '_{critic}} = \zeta {\theta _{critic}} + (1 - \zeta ){\theta '_{critic}}$, where $\zeta$ is set as 0.01.
We update Co-SA parameters with the Adam optimizer and adjust the learning rate based on valid loss, multiplying it by 0.95 with the patience of 50 epochs. Co-SA has trained 500 epochs for each task on 2080Ti devices.

\subsection{Ablation Study}
\par{In this section, we evaluate each component of the Co-SA framework using the MOSI and IEMOCAP databases, which are representative datasets for the MSA and MER tasks, respectively. Beyond presenting quantitative results, we also provide and compare various visualized outcomes to offer a comprehensive analysis of Co-SA. In the subsequent sections, we delve into the investigation of the MSD and DPSR modules within the SAE phase and the SAC phase, respectively.}

\begin{figure}[t]
  \centering
    \includegraphics[width=1.0\linewidth]{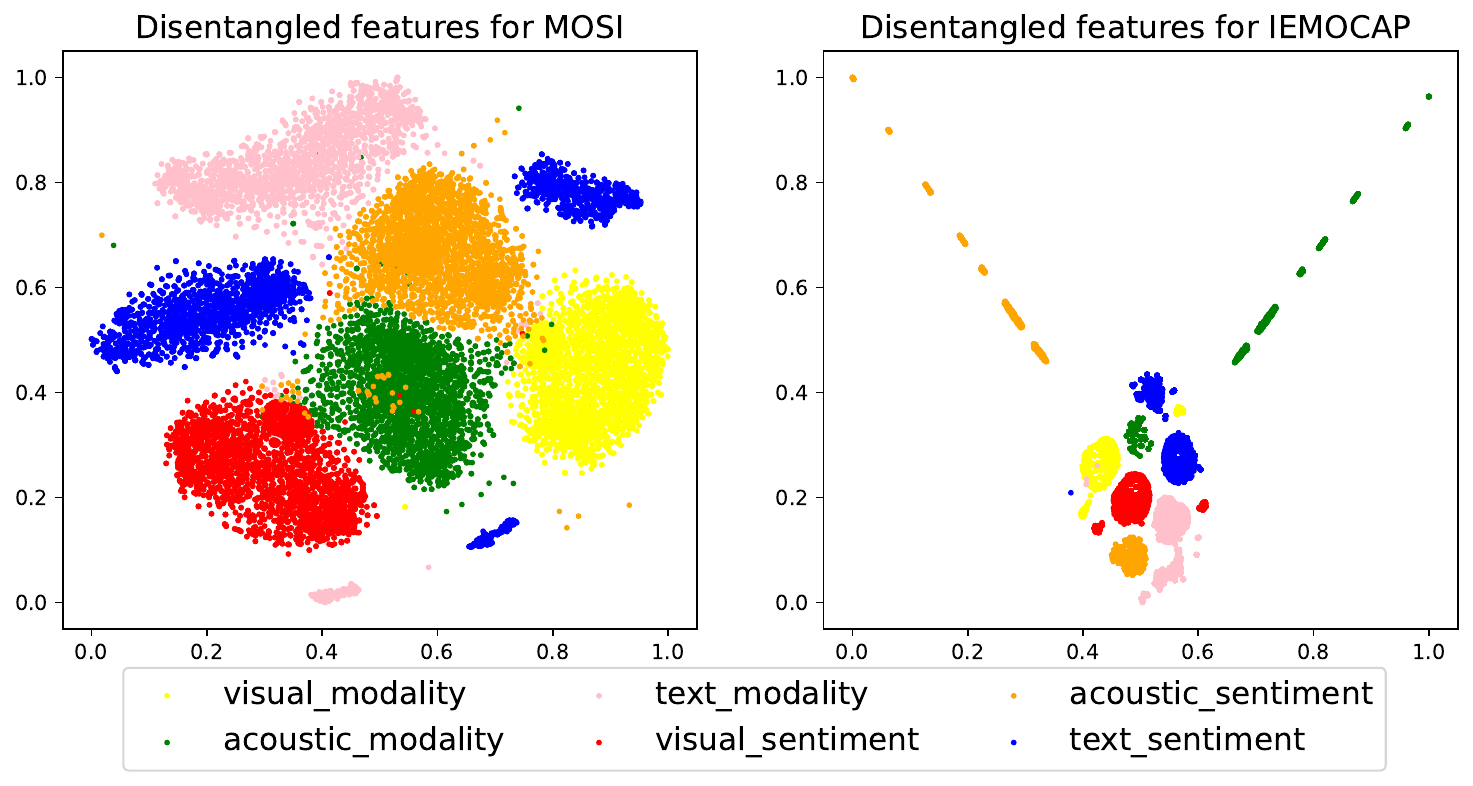}
    \caption{Disentangled features visualization on two databases.}
    \label{fig:msd}
\end{figure}

\subsubsection{Evaluation on the SAE Phase}
\par{The MSD module separates each modal signal into sentiment-related and modality-related properties. It involves three core operations corresponding with the reconstruction loss $L_r$, the modality loss $L_m$, and the sentiment-modality constraint loss $L_c$, respectively. Table.~\ref{tab:ablation1} presents comprehensive results for the MSD module, exploring the effectiveness of each loss. Notably, we define the model that excludes all proposed modules as the baseline method in Table.~\ref{tab:ablation1}. Compared with the baseline model, the MSD module efficiently improves model performance quantitatively on two tasks. Results, eliminating three loss functions by turns, indicate that the MSD module discounts to different extents when lacking one of the three loss functions. The compared results have demonstrated that Co-SA significantly benefits from the disentangled and homogenous sentiment properties in the MSD module.}

\par{Fig.~\ref{fig:msd} also presents intuitive disentangled features for three modal signals. In Fig.~\ref{fig:msd}, each sample is divided into modality and sentiment features. Sentiment features from the same modality are clustered together, as are modality features. However, sentiment clusters maintain a certain distance from their corresponding modality clusters. These visualized results suggest that Co-SA has successfully separated sentiment features from raw signals, thereby reducing the disturbances of diverse modal properties in subsequent steps. Moreover, sentiment features are not clustered across modalities, as their sequential variations need to be processed prior to fusion.}

\begin{figure}[t]
	\centering  
	\subfigbottomskip=2pt
	\subfigcapskip=-5pt
	\subfigure[Visualization of representations' similarities for the MOSI database.]{	\includegraphics[width=1.0\linewidth]{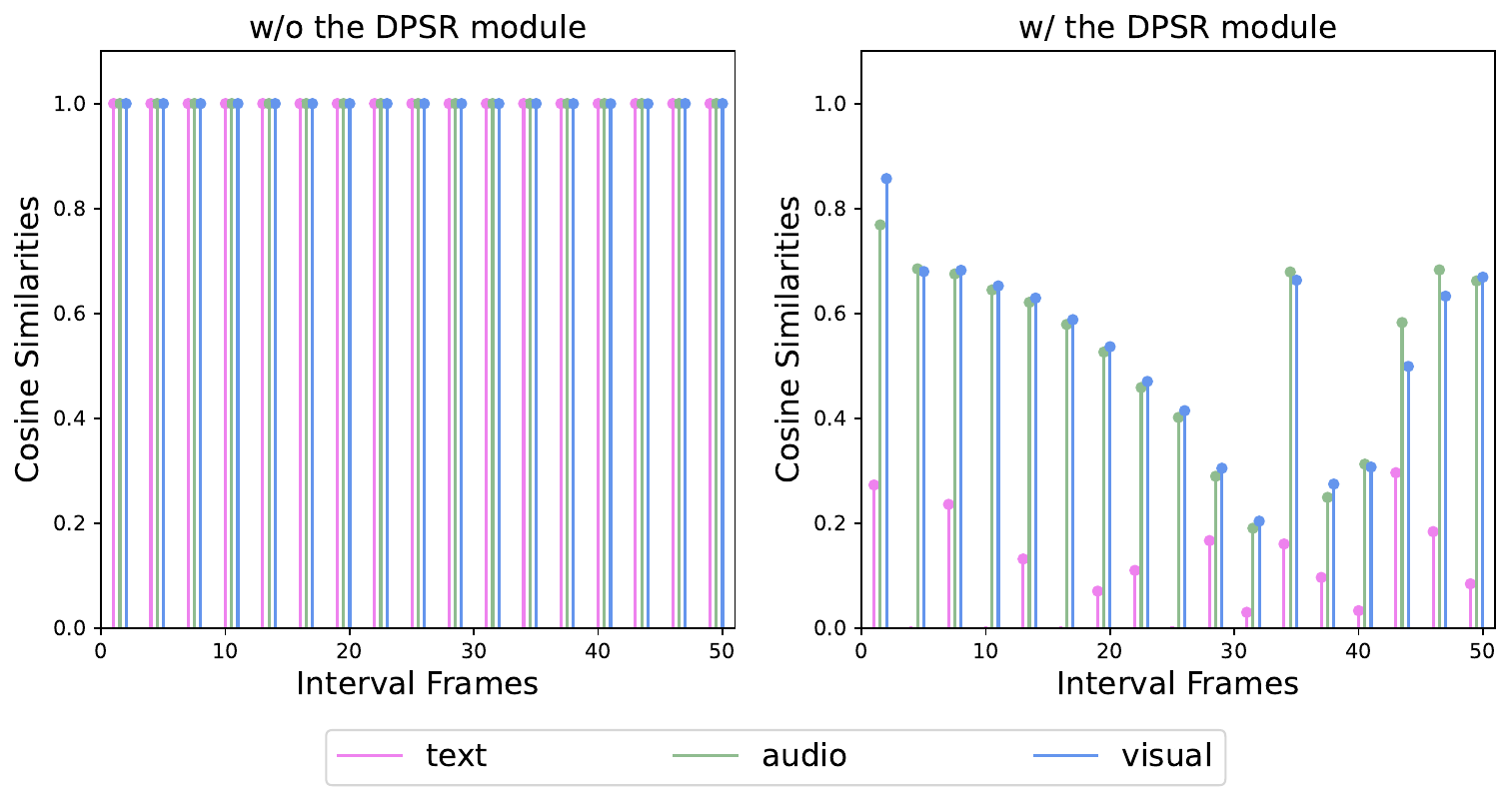}}
	\subfigure[Visualization of representations' similarities for the IEMOCAP database.]{	\includegraphics[width=1.0\linewidth]{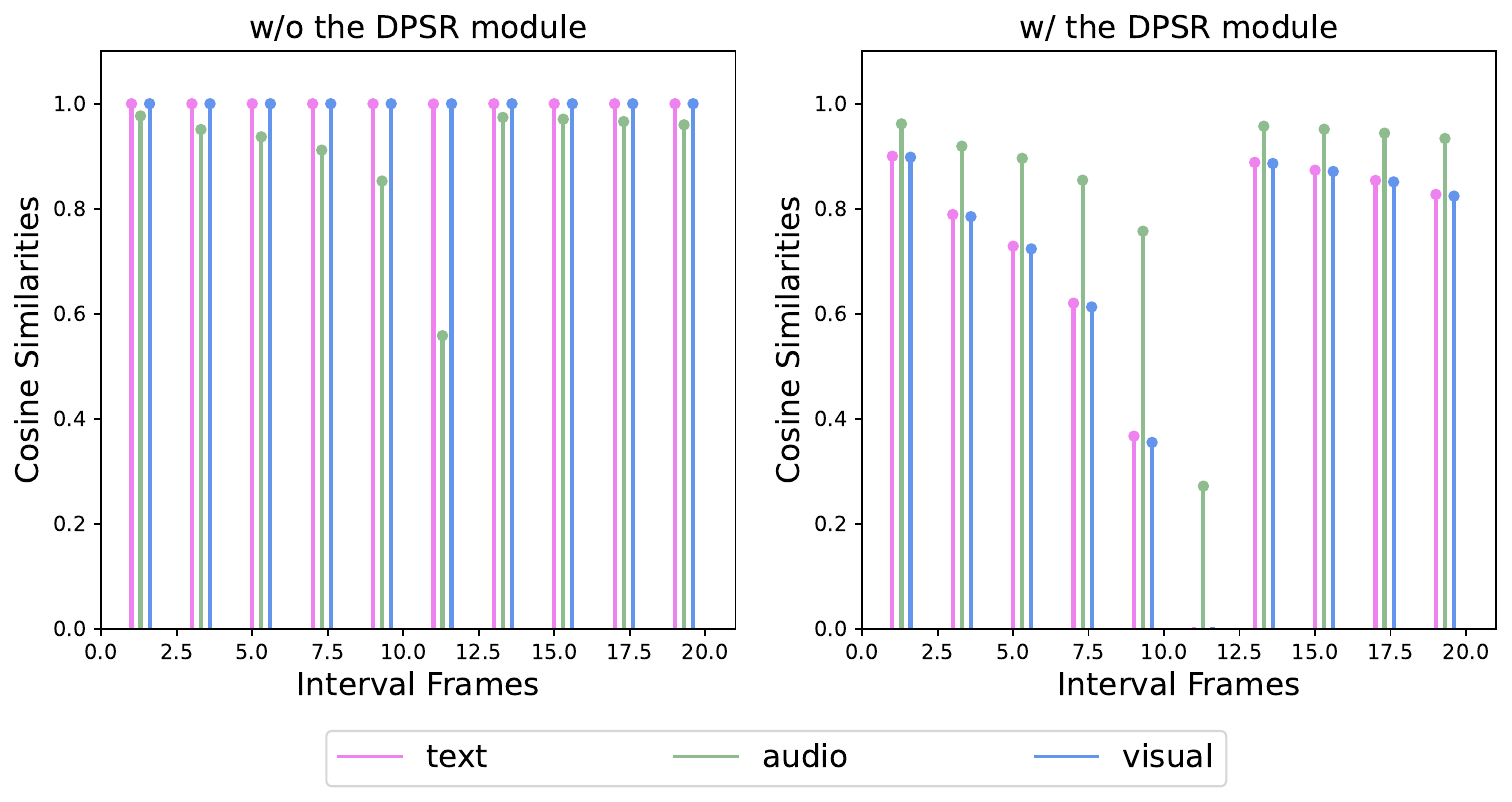}}
	\caption{Comparisons between representations' similarities on two databases.}
 \label{fig:psr}
\end{figure}

\par{The 'only DPSR' configuration in Table.~\ref{tab:ablation1} presents the ablation study results for the DPSR module. This module mitigates redundancy across representations and emphasizes sentiment variations over time. Compared to the baseline model, the DPSR module generally enhances all metrics on both tasks. Furthermore, the DPSR module utilizes the parameter $\eta$ to ease the constraint on frame representations as the time interval between frames expands. The performance of the DPSR module sees a slight decline when $\eta$ is removed, suggesting that repeated events may occur in certain sequences.}

\begin{table*}[t]
    \begin{threeparttable}
    \centering
    \tabcolsep=0.3cm
    \setlength\extrarowheight{3pt}
    \caption{
    Comparison with SOTA methods on the MOSI and MOSEI databases.}
    \begin{tabular}{ccccccccccc}
    \toprule[2pt]
                &\multicolumn{5}{c}{Dataset: MOSI} & \multicolumn{5}{c}{Dataset: MOSEI}\\
    \cmidrule(r){2-6} \cmidrule(r){7-11}
               & Acc7 ($\uparrow$) & Acc2 ($\uparrow$) & F1 ($\uparrow$)   & MAE ($\downarrow$)   & Corr ($\uparrow$) & Acc7 ($\uparrow$) & Acc2 ($\uparrow$) & F1 ($\uparrow$) & MAE ($\downarrow$) & Corr ($\uparrow$) \\
    \midrule[1pt]
    TFN-Glove~\cite{zadeh2017tensor}  & 32.2 & 76.4 & 76.3 & 1.017 & 0.604 & 49.8 & 79.4 & 79.7 & 0.610 & 0.671 \\
    TFN-BERT~\cite{zadeh2017tensor}   & 44.7 & 82.6 & 82.6 & 0.761 & 0.789 & 51.8 & 84.5 & 84.5 & 0.622 & 0.781 \\
    LMF-Glove~\cite{liu2018efficient}  & 30.6 & 73.8 & 73.7 & 1.026 & 0.602 & 50.0 & 80.6 & 81.0 & 0.608 & 0.677 \\
    LMF-BERT~\cite{liu2018efficient}   & 45.1 & 84.0 & 84.0 & 0.742 & 0.785 & 51.2 & 84.2 & 84.3 & 0.612 & 0.779 \\
    MFN-Glove~\cite{zadeh2018memory}  & 32.1 & 78.0 & 76.0 & 1.010 & 0.635 & 49.1 & 79.6 & 80.6 & 0.618 & 0.670 \\
    MFN-BERT~\cite{zadeh2018memory}   & 44.1 & 83.5 & 83.5 & 0.759 & 0.786 & 52.6 & 84.8 & 84.8 & 0.607 & 0.771 \\
    Mult-Glove~\cite{tsai2019multimodal} & 33.6 & 79.3 & 78.3 & 1.009 & 0.667 & 48.2 & 80.2 & 80.5 & 0.638 & 0.659 \\
    Mult-BERT~\cite{tsai2019multimodal}  & 41.5 & 83.7 & 83.7 & 0.767 & 0.799 & 50.7 & 84.7 & 84.6 & 0.625 & 0.775 \\
    QMF-Glove~\cite{li2021quantum}  & 35.5 & 79.7 & 79.6 & 0.915 & 0.696 & 47.9 & 80.7 & 79.8 & 0.640 & 0.658 \\
    GFN-BERT~\cite{mai2020modality}   & 47.0 & 84.3 & 84.3 & 0.736 & 0.790 & 51.8 & 85.0 & 85.0 & 0.611 & 0.774 \\
    ICCN-BERT~\cite{sun2020learning}  & 39.0 & 83.0 & 83.0 & 0.860 & 0.710 & 51.6 & 84.2 & 84.2 & \underline{\textbf{0.565}} & 0.713 \\
    RAVEN-Glove~\cite{wang2019words}& 33.8 & 78.8 & 76.9 & 0.968 & 0.667 & 50.2 & 79.0 & 79.4 & 0.605 & 0.680 \\ 
    MAG-BERT~\cite{rahman2020integrating}   & 42.9 & 83.5 & 83.5 & 0.790 & 0.769 & 51.9 & 85.0 & 85.0 & 0.602 & 0.778 \\
    TFR-Net~\cite{yuan2021transformer}    & 42.6 & 84.0 & 83.9 & 0.787 & 0.788 & 51.7 & 85.2 & 85.1 & 0.606 & 0.781 \\
    HyCon~\cite{mai2022hybrid}      & 46.6 & 85.2 & 85.1 & 0.713 & 0.790 & 52.8 & 85.4 & 85.6 & 0.601 & 0.776 \\
    MCL~\cite{mai2023excavating}        & 49.2 & 86.1 & 86.1 & 0.713 & 0.793 & 53.3 & 86.2 & 86.2 & 0.581 & \underline{\textbf{0.791}} \\
    Co-SA(add)  & \underline{\textbf{49.8}} & \underline{\textbf{87.2}} & \underline{\textbf{87.1}} & \underline{\textbf{0.685}} & \underline{0.813} & \underline{\textbf{54.5}} & \underline{\textbf{86.8}} & \underline{\textbf{86.7}} & \underline{0.575} & \underline{\textbf{0.791}}\\
    Co-SA(concatenate) & \underline{49.5} & \underline{86.6} & \underline{86.5} & \underline{0.689} & \underline{\textbf{0.814}} &  \underline{54.0} & \underline{86.4} & \underline{86.4} & 0.577 & \underline{0.790}\\

    \bottomrule[2pt]
    \end{tabular}
    \label{tab:mosi}
    \begin{tablenotes}
    \scriptsize
    \item[] The optimal results are bolded and underlined, and the sub-optimal results are underlined.
    \end{tablenotes}
    \end{threeparttable}
\end{table*}

\par{To further scrutinize the DPSR module, we compute the average similarity between features across various time intervals. Fig.~\ref{fig:psr} illustrates the similarity comparisons between representations without (w/o) and with (w/) the DPSR module. Representations, when not modulated by the DPSR module, show considerable similarities across different times since they are defined by objective characteristics, such as visual AUs and acoustic MFCCs. These objective characteristics exhibit minimal variation within the same sequence and struggle to reflect sentiment fluctuations. However, when the DPSR module is employed, similarities among frames significantly decrease as frame intervals increase. Interestingly, the similarities of representations tend to increase again when their time interval surpasses a certain threshold. This phenomenon fortuitously confirms our assumption in the quantitative ablation study concerning the DPSR module.}

\begin{figure}[t]
  \centering
    \includegraphics[width=1.0\linewidth]{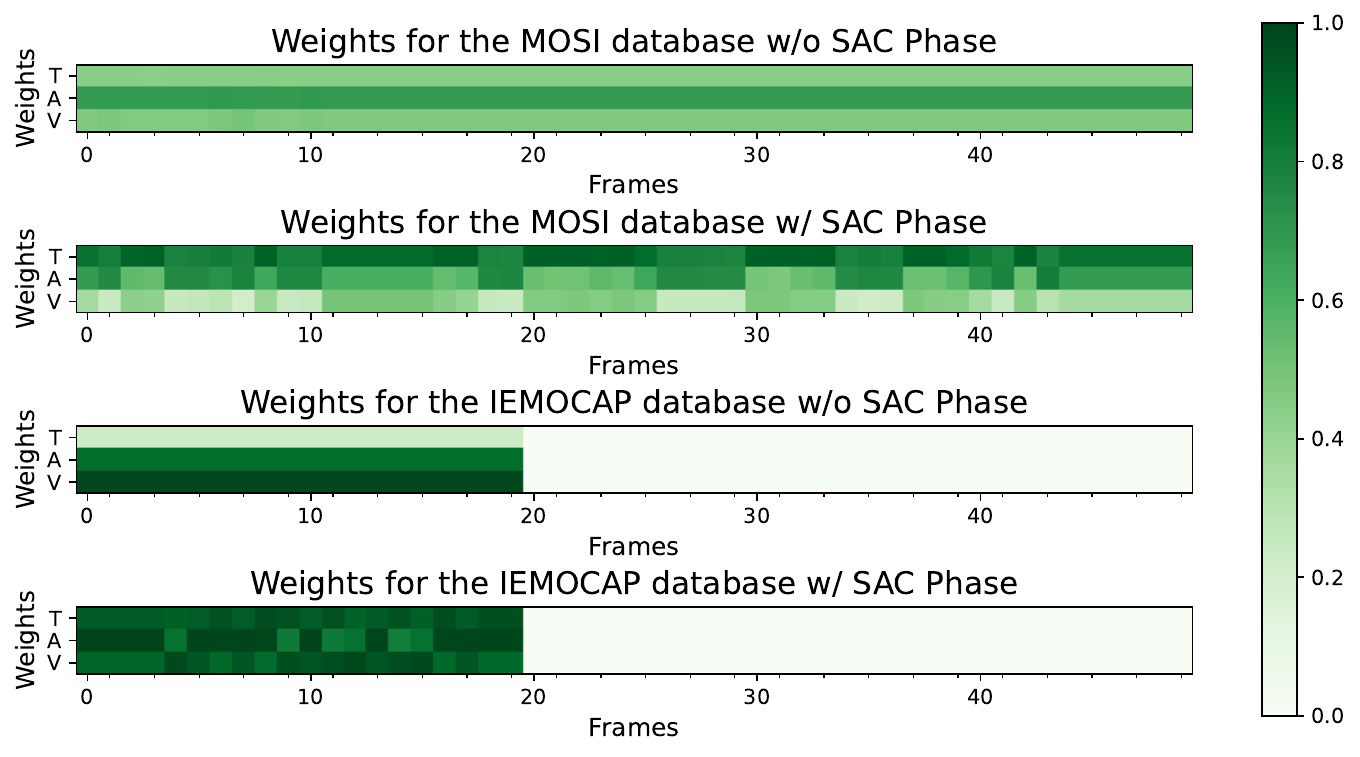}
    \caption{Weight comparisons w/o and w/ the SAC phase on two databases.}
    \label{fig:sac}
\end{figure}

\begin{figure}[h]
\centering  
\subfigure[MAE results on the MOSI database under different hyper-parameters.]{\includegraphics[width=0.9\linewidth]{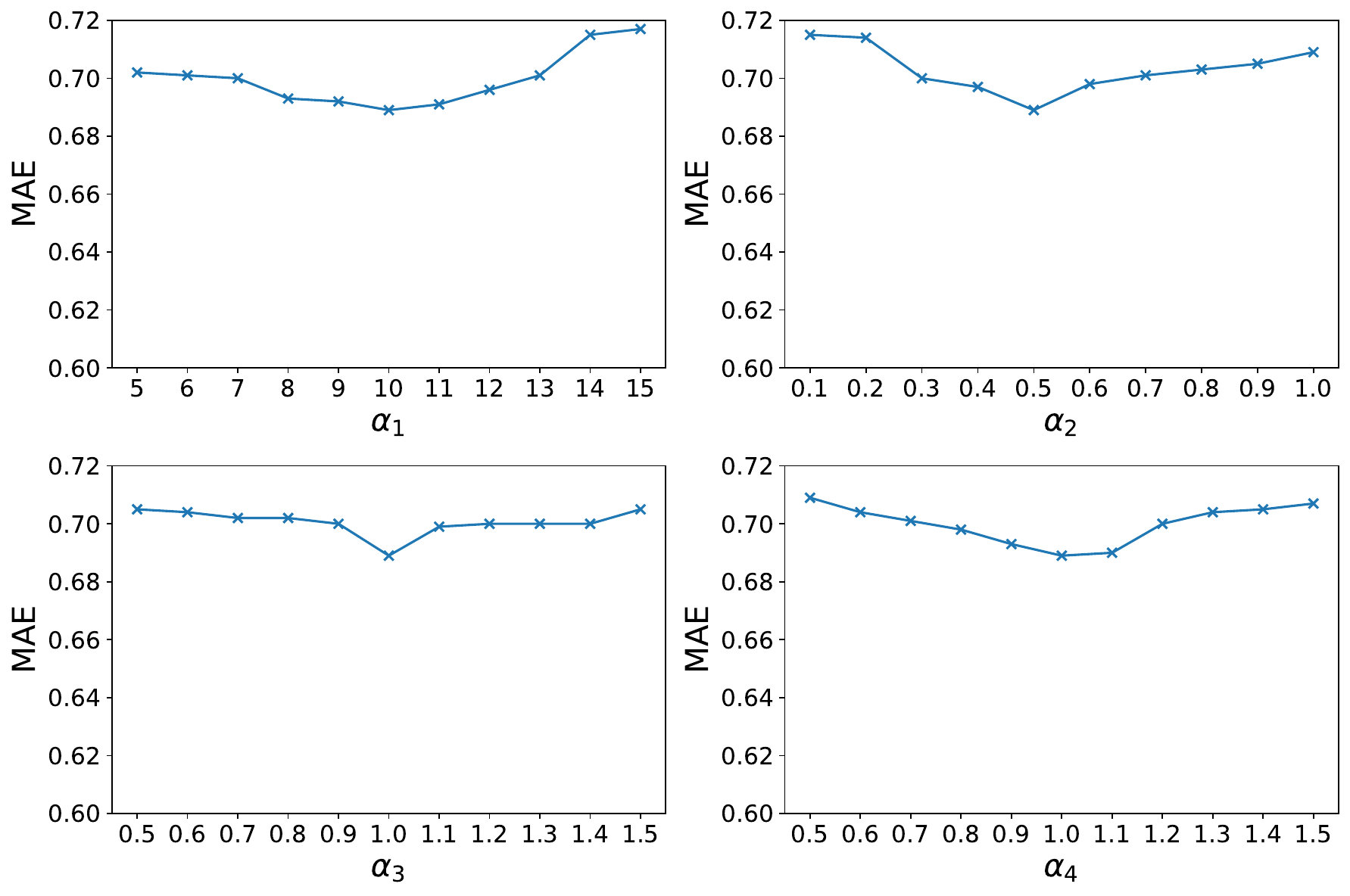}}
\centering
\subfigure[2-class accuracies on the IEMOCAP database under different hyper-parameters.]{\includegraphics[width=0.9\linewidth]{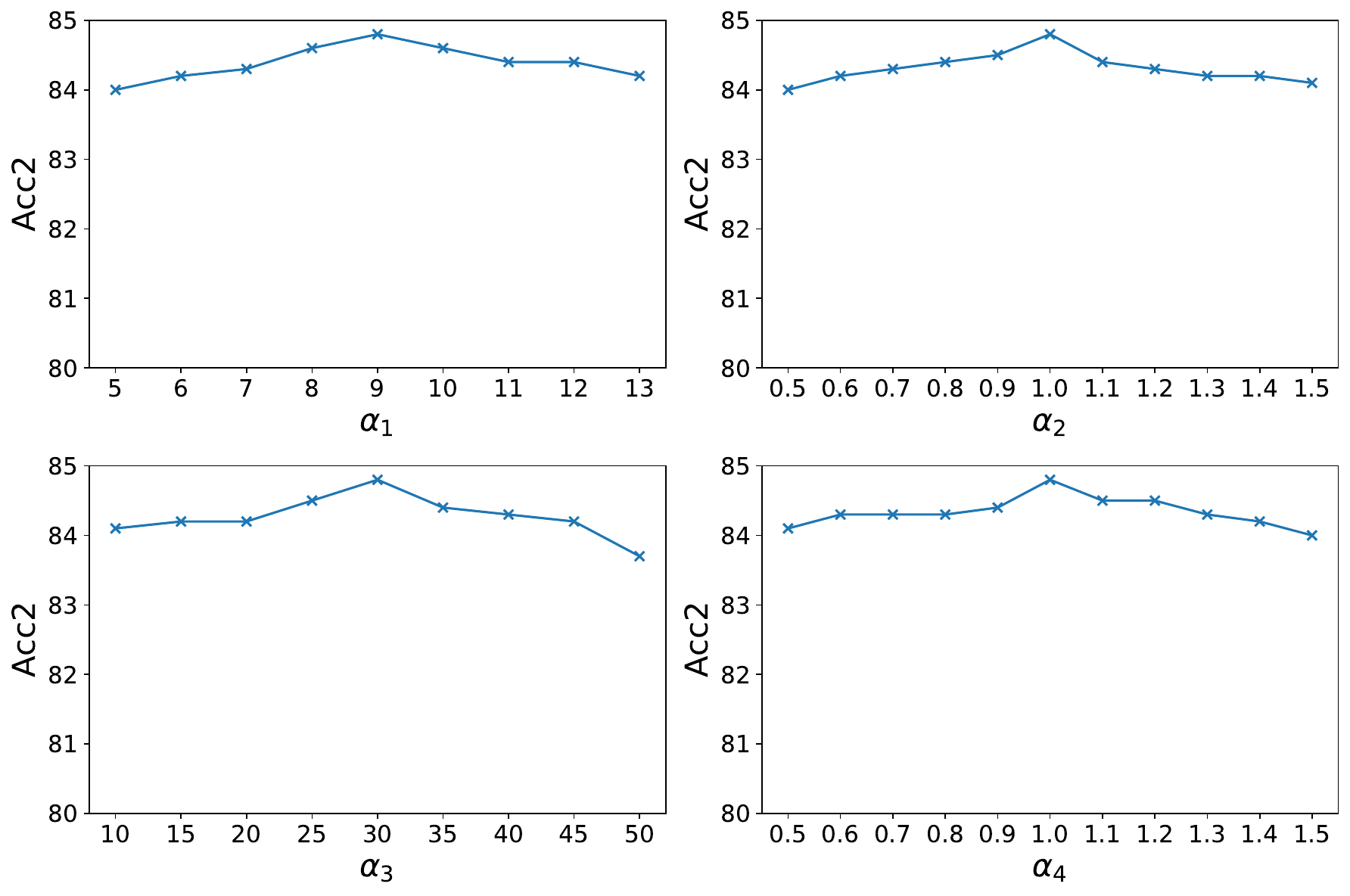}}

\caption{Ablation studies on hyper-parameters. $\alpha_1$, $\alpha_2$, $\alpha_3$, and $\alpha_4$ are the weights for the prediction, MSD, DPSR, and SAC modules, respectively.}
\label{fig:weight}
\end{figure}

\subsubsection{Evaluation on the SAC Phase}
\par{Table.~\ref{tab:ablation1} also presents a comparison between models without (w/o) and with (w/) the SAC phase. In the baseline model, we eliminate the cooperative sentiment agents and learn weights for the fused features directly. Compared to the baseline model, the SAC module consistently enhances the model's performance on both tasks, owed largely to its superior weight-learning mechanism. However, due to the constraints observed in the representation of sentiment agents, the model that exclusively employs the SAC module does not achieve optimal results. 
To provide a more intuitive understanding of the SAC module, we visualize the learned weights $w_i$ in Fig.~\ref{fig:sac}. Each value in $w_i$ ranges from 0 to 1, signifying the importance of the corresponding features. Notably, the data in the MOSI and IEMOCAP databases are aligned across modalities. Co-SA captures modality-shared features when it learns larger weights at the same frame for all three modalities. Conversely, it captures complementary features when larger weights are distributed among different frames. Initially, models that directly learn fused properties tend to rely on one or two modalities, thereby failing to capture diverse cross-modal features. For instance, the model heavily depends on acoustic signals in the MSA task and emphasizes features in the acoustic and visual modalities in the MER task. This over-reliance on specific modalities overlooks the diversity of multimodal features, rendering the model less robust. In contrast, the proposed SAC phase, benefiting from the feedback reward, repeatedly learns the fused representations and is adept at mining valuable properties across all modalities. Furthermore, compared to conventional methods, Co-SA has unearthed diverse inter-modal features, encompassing both common and complementary aspects.}

\par{\subsubsection{Evaluation on Hyper-parameters} The objective function of Co-SA is composed of the prediction loss $L_p$, the loss $L_{msd}$ for the MSD module,  the loss $L_{dpsr}$ for the DPSR module, and the loss $L_{actor} + L_{critic}$ for the SAC phase, with weights of $\alpha_1$, $\alpha_2$, $\alpha_3$, and $\alpha_4$. To establish a more rigorous experimental setup, we conduct extensive experiments over a wide range of values for these four weights. Fig.~\ref{fig:weight} illustrates the weights and corresponding evaluation metrics for the MSA and MER tasks. Co-SA achieves the optimal MAE value for the MSA task when setting $\alpha_1$, $\alpha_2$, $\alpha_3$, and $\alpha_4$ as 10, 0.5, 1.0, and 1.0, respectively. For the MER task, it performs best at the setting of $\alpha_1 = 9$, $\alpha_2 = 1.0$, $\alpha_3 = 30$, and $\alpha_4 = 1.0$. Furthermore, the performance of Co-SA exhibits minor fluctuations when exploring the optimal weights within a certain range, demonstrating its robustness.
}

\begin{table*}[t]    
    \begin{threeparttable}
    \centering
    \tabcolsep=0.5cm
    \setlength\extrarowheight{3pt}
    \caption{Comparison with SOTA methods on the IEMOCAP database.}
    \begin{tabular}{ccccccccccc}
    \toprule[2pt]
                & \multicolumn{2}{c}{Happy} & \multicolumn{2}{c}{Sad} & \multicolumn{2}{c}{Angry} & \multicolumn{2}{c}{Neutral} & \multicolumn{2}{c}{Average} \\
\cmidrule(r){2-3} \cmidrule(r){4-5} \cmidrule(r){6-7} \cmidrule(r){8-9} \cmidrule(r){10-11}
                & Acc  & F1 & Acc & F1 & Acc & F1 & Acc & F1 & Acc & F1 \\
    \midrule[1pt]
    MFN~\cite{zadeh2018memory}        & 86.5 & 84.0 & 83.5 & 82.1 & 85.0 & 83.7 & 69.6 & 69.2 & 81.2 & 79.8 \\
    Graph-MFN~\cite{zadeh2018memory}  & 86.8 & 84.2 & 83.8 & 83.0 & 85.8 & 85.5 & 69.4 & 68.9 & 81.5 & 80.4 \\
    RAVEN~\cite{wang2019words}      & 87.3 & 85.8 & 83.4 & 83.1 & 87.3 & 86.7 & 69.7 & 69.3 & 81.9 & 81.2 \\
    LMF~\cite{liu2018efficient}        & 86.9 & 82.3 & 85.4 & 84.7 & 87.1 & 86.8 & 71.6 & 71.4 & 82.8 & 81.3 \\
    MulT~\cite{tsai2019multimodal}       & 87.4 & 84.1 & 84.2 & 83.1 & 88.0 & 87.5 & 69.9 & 68.4 & 82.4 & 80.8 \\
    HFFN~\cite{mai2019divide}       & 86.8 & 82.1 & 84.4 & 84.5 & 86.6 & 85.8 & 69.6 & 69.3 & 81.9 & 80.4 \\
    TCM-LSTM~\cite{mai2021analyzing}   & 87.2 & 84.8 & 84.4 & 84.9 & 89.0 & 88.6 & 71.3 & 71.2 & 83.0 & 82.4 \\
    HyCon~\cite{mai2022hybrid}      & 88.0 & 85.5 & 86.2 & 85.9 & \underline{89.4} & 89.2 & 70.4 & 70.5 & 83.5 & 82.8 \\
    MCL~\cite{mai2023excavating}        & \underline{\textbf{88.8}} & \underline{\textbf{86.8}} & 86.6 & 86.6 & \underline{\textbf{90.3}} & \underline{\textbf{90.3}} & 71.6 & 71.4 & 84.3 & 83.8 \\
    Co-SA(add)  & \underline{88.2} & \underline{\textbf{86.8}} & \underline{87.2} & \underline{\textbf{87.1}} & \underline{\textbf{90.3}} & \underline{90.1} & \underline{73.4} & \underline{73.1} & \underline{\textbf{84.8}} & \underline{\textbf{84.3}}\\
    Co-SA(concatenate)& 88.1 & \underline{86.3} & \underline{\textbf{87.6}} & \underline{87.0} & 89.1 &  89.0 & \underline{\textbf{74.0}} & \underline{ \textbf{73.5}} & \underline{84.7} & \underline{83.9}\\
    \bottomrule[2pt]
    \end{tabular}
    \label{tab:iemocap}

    \begin{tablenotes}
    \scriptsize
    \item[] The optimal results are bolded and underlined, and the sub-optimal results are underlined.
    \end{tablenotes}
    \end{threeparttable}
\end{table*}

\subsection{Comparison With State-of-the-Art Works}
\par{In this section, we compare Co-SA with several state-of-the-art methods. Co-SA produces unimodal weights that dictate the features used for fusion. Without loss of generality, we employ two classic fusion modes to obtain a joint multimodal representation: addition and concatenation operations, denoted as Co-SA(add) and Co-SA(concatenate), respectively. Comparative results on the MOSI and MOSEI databases are presented in Table~\ref{tab:mosi}.
Among the compared works, MCL has so far achieved the best results on both databases. MCL is also an approach that aims to learn high-quality multimodal representations by establishing correlations between modalities. Compared with MCL, Co-SA(add) enhances performance by 0.6\%, 1.1\%, 1.0\%, 0.028, and 0.02 on Acc7, Acc2, F1 score, MAE, and Corr, respectively, on the MOSI database. On the MOSEI database, it improves performance by 1.2\%, 0.6\%, and 0.5\% on Acc7, Acc2, and F1 score, respectively. The correlation (Corr) predicted by Co-SA(add) for the MOSEI database remains on par with MCL. Meanwhile, the MAE of Co-SA for the MOSEI database is slightly worse than ICCN-BERT. This generalized improvement on the MSA task is attributed to the well-designed joint representation learning process.
Regarding the concatenation fusion mode, Co-SA enhances Acc7, Acc2, F1 score, MAE, and Corr by 0.3\%, 0.5\%, 0.4\%, 0.24, and 0.21, respectively, on the MOSI database, and boosts 0.7\%, 0.2\%, and 0.2\% on Acc7, Acc2, and F1 score, respectively, on the MOSEI database. The correlation between Co-SA's predictions and the ground truth experiences a 0.01 drop compared with MCL. However, the overall results of both fusion modes demonstrate a stable improvement compared with state-of-the-art works.}

\par{Table~\ref{tab:iemocap} presents a comparison with several state-of-the-art (SOTA) works on the IEMOCAP database. Similar to the MSA task, MCL has achieved the best results for the MER task. Compared with MCL, Co-SA exhibits slightly inferior performance in the happy class but significantly improves the recognition rates in the neutral class. Specifically, Co-SA enhances the accuracy of the neutral emotion by 1.8\% and 2.4\%, respectively, when using the two fusion modes. Moreover, Co-SA also significantly improves the mean accuracy across the four emotions, outperforming all existing works.}

\subsection{Evaluation on Modalities}

\begin{table}[]
    \begin{threeparttable}
    \centering
    \tabcolsep=0.18cm
    \setlength\extrarowheight{7pt}
    \caption{Performance of arbitrary combinations of modalities.}
    \label{tab:modality}
    \begin{tabular}{cccccc}
    \toprule[2pt]
          & Acc7 ($\uparrow$)  & Acc2 ($\uparrow$) & F1 ($\uparrow$)   & MAE ($\downarrow$)    & Corr ($\uparrow$) \\
    \midrule[1pt]
    V     & 21.0 & 57.7 & 55.7 & 1.447  & 0.093 \\
    A     & 20.2 & 53.3 & 52.7 & 1.448  & 0.132 \\
    T     & \underline{\underline{45.4}} & \underline{\underline{84.1}} & \underline{\underline{84.0}} & \underline{\underline{0.736}}  & \underline{\underline{0.792}} \\
    V+A   & 22.9 & 57.4 & 57.2 & 1.442  & 0.135 \\
    V+T   & 49.3 & 85.3 & 85.3 & 0.703  & 0.804\\
    A+T   & 47.7 & 85.8 & 85.7 & 0.714  & 0.798 \\
    V+A+T & \underline{\underline{49.8}} & \underline{\underline{87.2}} & \underline{\underline{87.1}} & \underline{\underline{0.685}}  & \underline{\underline{0.813}} \\
    \bottomrule[2pt]
    \end{tabular}

    \begin{tablenotes}
    \scriptsize
    \item[] Significant results are underlined.
    \end{tablenotes}
    \end{threeparttable}
\end{table}

\par{In this section, we delve into the significance of the three modalities. Taking the MOSI database as an example, we conduct experiments with various combinations of modalities and present the results in Table~\ref{tab:modality}. In Table~\ref{tab:modality}, "V", "A", and "T" denote the visual, acoustic, and text modalities, respectively.
From the comparison among individual modalities, it is apparent that text data plays a pivotal role among the three modalities. This phenomenon is largely due to the fact that both the visual and acoustic modalities are constrained to use extracted features (such as AUs and MFCCs) as input because of privacy concerns, whereas raw text data can be directly incorporated into Co-SA. Raw data offer a far richer source of information than extracted features.
The amalgamation of text with both visual and acoustic modalities yields a noticeable improvement across all metrics, indicating that visual and acoustic signals provide effective supplementary or complementary information. Ultimately, Co-SA achieves optimal performance when all modal signals are engaged. }

\section{Conclusion}
In this paper, we aim to harness effective unimodal properties to construct a comprehensive joint multimodal representation. We analyze that existing works devote more to modality-fusion strategies. However, some pre-defined modality-fusion mechanisms limit models in perceiving diverse cross-modal features. Inspired by cooperative control, we establish distinct sentiment agents for multimodal signals, each comprising the MSD and DSPR modules that capture sentiment variations. Subsequently, these sentiment agents interact via a policy-learning and policy-optimization mechanism to capture significant cross-modal features. Quantitative results from two tasks have validated the effectiveness of our proposed modules. Moreover, the visualization of intermediate features provides a clear and intuitive understanding of each component of Co-SA. The introduction of multiple sentiment agents opens up a new avenue for research in multimodal data analysis.

\bibliographystyle{IEEEtran}
\bibliography{Co-SA}

\end{document}